\pdfoutput=1

\documentclass[11pt]{article}

\usepackage[]{EACL2023}

\usepackage{times}
\usepackage{latexsym}

\usepackage[T1]{fontenc}

\usepackage[utf8]{inputenc}

\usepackage{microtype}

\usepackage{inconsolata}
\usepackage{hyperref}       
\usepackage{url}            
\usepackage{booktabs}       
\usepackage{amsfonts}       
\usepackage{nicefrac}       
\usepackage{microtype}      
\usepackage{xcolor}         
\usepackage{times}
\usepackage{multirow}
\usepackage{latexsym}
\usepackage{hyperref}
\usepackage{microtype}
\usepackage{graphicx}
\usepackage[T1]{fontenc}
\usepackage{times}
\usepackage{float}
\usepackage{microtype}
\usepackage{graphicx}
\usepackage{ulem}
\usepackage{comment}
\usepackage{algpseudocode}
\usepackage{tensor}
\usepackage{stfloats}
\usepackage{amssymb}
\usepackage{pdfpages}
\usepackage{latexsym}
\usepackage{amsmath, xparse}

\usepackage[utf8]{inputenc}
\usepackage{booktabs}
\usepackage{tabularx}
\usepackage{amstext}

\title{KronA: Parameter Efficient Tuning with \underline{Kron}ecker \underline{A}dapter}

\author{Ali Edalati \\ McGill University \\  \small{ali.edalati@mail.mcgill.ca} \And Marzieh Tahaei \\ Huawei Noah's Ark Lab \\ \small{marzieh.tahaei@huawei.com} 
\And Ivan Kobyzev \\ Huawei Noah's Ark Lab \\ \small{ivan.kobyzev@huawei.com}
\AND Vahid Partovi Nia \\ Huawei Noah's Ark Lab \\ \small{vahid.partovinia@huawei.com} \And James J. Clark \\ McGill University \\ \small{james.clark1@mcgill.ca} \And  Mehdi Rezagholizadeh \\ Huawei Noah's Ark Lab \\ \small{mehdi.rezagholizadeh@huawei.com}} 

\begin{document}
\maketitle
\begin{abstract}
Fine-tuning a Pre-trained Language Model (PLM) on a specific downstream task has been a well-known paradigm in Natural Language Processing. However, with the ever-growing size of PLMs, training the entire model on several downstream tasks becomes very expensive and resource-hungry. Recently, different Parameter Efficient Tuning (PET) techniques are proposed to improve the efficiency of fine-tuning PLMs. One popular category of PET methods is the low-rank adaptation methods which insert learnable truncated SVD modules into the original model either sequentially or in parallel. However, low-rank decomposition suffers from limited representation power. In this work, we address this problem using the Kronecker product instead of the low-rank representation. We introduce KronA, a Kronecker product-based adapter module for efficient fine-tuning of Transformer-based PLMs. We apply the proposed methods for fine-tuning T5 on the GLUE benchmark to show that incorporating the Kronecker-based modules can outperform state-of-the-art PET methods.
\end{abstract}

\section{Introduction}
Large PLMs are used as a backbone model in a variety of NLP tasks to achieve state-of-the-art results \cite{devlin-etal-2019-bert,radford2019language}.  These large pre-trained models are adapted to the downstream applications either via in-context learning or fine-tuning of the model parameters. In-context learning imposes substantial
memory and computational overhead during inference as all the training examples have to be processed for each sample \cite{liu2022few}. On the other hand, full Fine-Tuning (FT) the entire model provides both less inference latency and improved accuracy. However, as these models become larger, full fine-tuning of their parameters becomes more challenging. Additionally, one has to store an entire model checkpoint for each downstream application, which makes deployment and switching between different tasks extremely inefficient. 

To address these challenges, several works have proposed to insert a small number of trainable parameters while freezing most (or even all) of the pre-trained model parameters. 
This significantly reduces the memory and computation requirements for fine-tuning. Furthermore, instead of storing one copy of the entire model, a small set of tuned parameters can be stored for each task. We refer to these methods as PET methods. 

Among the PET methods, soft prompts \cite{li-liang-2021-prefix,lester2021power} prepend trainable parameters to the input of the layers. The increase in length of the embedding layers leads to a significant computation overhead during the inference. 

In another category of the PET methods, adapter modules are inserted  \cite{pmlr-v97-houlsby19a,karimi2021compacter,he2022towards} into the Transformer. Adapters are low-rank modules that are composed of an up projection followed by a down projection. One limitation of these approaches is that they increase the computational overhead and the latency during the inference which makes them inefficient for latency-critical scenarios. 

Therefore, Low Rank Adaption (LoRA) \cite{hu2021lora} was developed, which also uses extra low-rank modules as the trainable parameters. However, once fine-tuned, the task-specific parameters can be merged with the original pre-trained model weights,  making the latency and energy requirements for inference, intact. Despite of fast inference, usually LoRA suffers from an accuracy drop compared to full fine-tuning. This is because of the strong assumption imposed by its low-rank structure for task-specific updates. 

Kronecker product decomposition is another factorization method that does not rely on the low-rank assumption. This powerful decomposition method, when used for model compression, has proven to outperform low-rank factorization methods \cite{thakker2019pushing,hameed2021convolutional}. It has also been successfully used for the compression of Transformer-based language models \cite{tahaei2021kroneckerbert,edalati2021kronecker}.

Inspired by the ubiquitous success of Kronecker decomposition, in this work we replace low-rank decomposition in LoRA with Kronecker product decomposition to develop the Kronecker Adapter (KronA). We show that this simple modification can improve the accuracy without increasing the inference latency. Also for applications where an increase in latency is tolerable, we propose to use $\text{KronA}^\text{B}$. This module is a version of KronA developed to be utilized in parallel to Feed-Forward Network (FFN) modules and achieves significant improvements over full fine-tuning on the General Language Understanding Evaluation (GLUE) benchmark \cite{wang-etal-2018-glue}. In addition, when a proposed learnable residual connection is added to $\text{KronA}^\text{B}$, $\text{KronA}^\text{B}_\text{res}$ is developed to achieve even better results.

We evaluated our methods on the GLUE benchmark \cite{wang-etal-2018-glue} to study the impact of Kronecker product on the performance. To summarize, our contributions are:
\begin{itemize}
    \item Proposing KronA, a \underline{Kron}ecker \underline{A}dapter module that can be inserted in parallel to the weight matrices and is suitable for latency-critical scenarios.
    \item Using the KronA module in parallel to the FFN module ($\text{KronA}^\text{B}$) along with a learnable residual connection ($\text{KronA}^\text{B}_\text{res}$) to further improve the accuracy at the cost of increased inference latency.
    \item Providing evaluation of the methods in comparison to the state-of-the-art in terms of the GLUE score, training time and inference latency.
\end{itemize}

\begin{figure*}
    \centering
    \includegraphics[width=\textwidth]{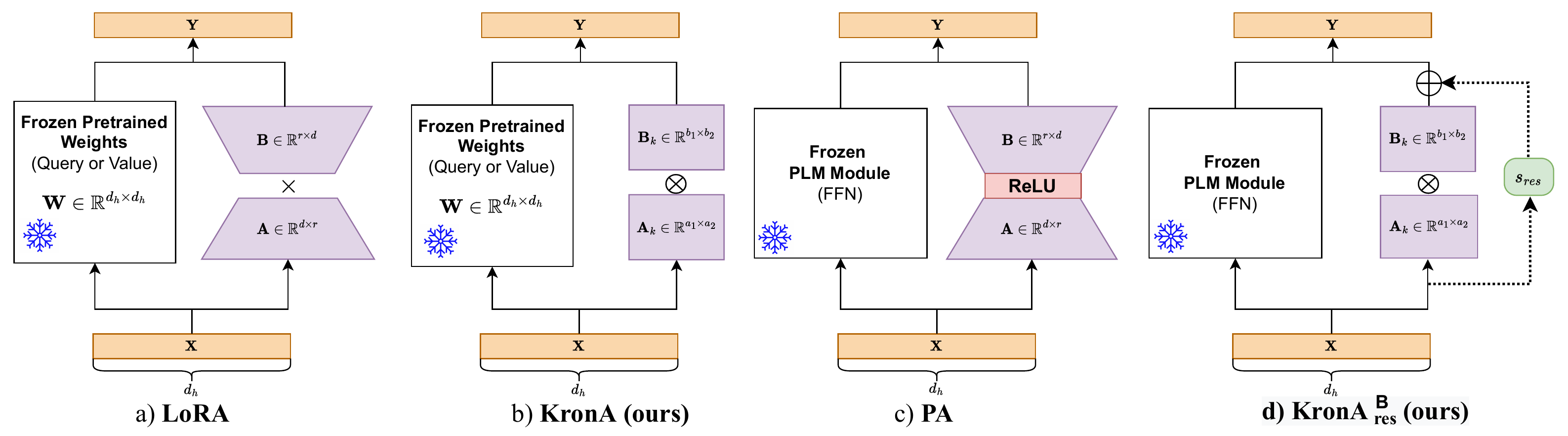}
    \caption{This figure shows the structure of the proposed Kronecker-based modules and their low-rank counterparts. For simplicity, the scaling factor at output of the modules is not depicted. Figure d shows $\text{KronA}^\text{B}_\text{res}$. The residual connection is depicted by a dotted-line to remind that this connection can simply be removed to have $\text{KronA}^\text{B}$.}
    \label{fig:1}
\end{figure*}

\section{Related Works and Baselines}
\citep{zaken2021bitfit} proposed freezing the weights and tuning only biases or a subset of biases in the PLM to fine-tune it on the downstream tasks. This technique, called BitFit, is parameter-efficient and fast, but it usually cannot achieve good performance compared to the state-of-the-art methods.

\cite{pmlr-v97-houlsby19a} introduced Adapters as a PET method. In this method, the entire parameters of a PLM are frozen and some trainable modules, called Adapters, are inserted after the FFN or attention blocks, sequentially. Every Adapter module has a down projection, a non-linear function and an up projection in addition to a residual connection. \cite{he2022towards} developed Parallel-Adapter (PA) which outperforms Adapter. PA is inserted in parallel to the original PLM module and has a scaling factor, but the residual connection inside the module is removed since the PLM module has a residual connection (Figure \ref{fig:1}.c). \cite{he2022towards} also introduced a unified view over PET methods and combined some of the techniques like Prefix tuning \cite{li-liang-2021-prefix} and PA.

In \cite{karimi2021compacter}, a modified version of the sequential adapter is used for PET, named Compacter. In Compacter, Kronecker product of multiple Kronecker factors are added to reconstruct the module's weight matrix. Each Kronecker factor itself is a result of matrix multiplication of two sub-factors. Compacter achieves good results on GLUE, but it is notably slow in both training and inference phases.

LoRA \cite{hu2021lora} inserts trainable modules which are made from a down projection and an up projection in different parts of a PLM. It is worth mentioning that the authors recommended to insert the LoRA modules in parallel to the query and value matrices.  During the training, the PLM weights are frozen and only the LoRA modules are tuned. During the inference, the LoRA weights are merged with the original weight matrices of the PLM. Therefore, in contrast to the Adapter and Compacter, LoRA does not increase the inference time. 

Since LoRA, PA and Compacter have outperformed most of the existing baselines such as Pfeiffer-Adapter \cite{pfeiffer2020adapterfusion}, AdapterDrop \cite{ruckle-etal-2021-adapterdrop}, VGLM-Adapter \cite{lin-etal-2020-exploring} Prompt tuning \citep{lester-etal-2021-power}, and Prefix tuning \cite{li-liang-2021-prefix}, we considered these works as baseline for comparison in this work.

\section{Methodology}
\label{sec:meth}
\subsection{Kronecker Product}
Kronecker product is an operation on two input matrices ($\mathbf{A} \in \mathbb{R}^{a_1 \times a_2}$ and $\mathbf{B} \in \mathbb{R}^{b_1 \times b_2}$) that results a block matrix ($\mathbf{W} \in \mathbb{R}^{w_1 \times w_2}$). In $\mathbf{W}$, each block $(i, j)$ is equal to the multiplication of the element $a_{i,j}$ and the matrix $\mathbf{B}$. Also, the shape of the resulting matrix is $(w_1, w_2)$, where $w_1 = a_1 \times b_1$ and $w_2 = a_2 \times b_2$. Equation \ref{eq:kron} shows how the Kronecker product works. for more details see  \cite{Henderson1983OnTH}.
\begin{equation}
    \mathbf{W} = \mathbf{A}\otimes\mathbf{B} = \begin{bmatrix}
  a_{11} \mathbf{B} & \cdots & a_{1n}\mathbf{B} \\
             \vdots & \ddots &           \vdots \\
  a_{m1} \mathbf{B} & \cdots & a_{mn} \mathbf{B}
\end{bmatrix}
\label{eq:kron}
\end{equation}

The Kronecker product has some interesting features that makes it suitable for PET. First, it is not rank deficient. In other words, unlike the low-rank down-projection used in LoRA and Adapter, Kronecker product decomposition maintains the rank of the input matrix. Second, to speed up and reduce the required number of FLOPS, one can avoid the reconstruction of $\mathbf{W}$ and use Equation \ref{eq:kronecker_equivalance} to calculate the output of a Kronecker module. More precisely, multiplying an input vector $\mathbf{x} \in \mathbb{R}^{d_h}$, where $d_h$ is the embedding dimension, by the Kronecker product of $\mathbf{A}$ and $\mathbf{B}$ can be performed using the following equation:
\vspace{-3pt}
\begin{equation}
    (\mathbf{A} \otimes \mathbf{B})\mathbf{x} = \gamma(\mathbf{B}\eta_{b_2\times a_2}(\mathbf{x})\mathbf{A^\top})
\label{eq:kronecker_equivalance}
\end{equation}
where  $\mathbf{A}^\top$ is $\mathbf{A}$ transpose. $\eta_{m \times n}(\mathbf{y})$ is a mathematical operation that reshapes a vector  $\mathbf{y} \in \mathbb{R}^{mn}$  into a matrix of size $m \times n$. $\gamma(\mathbf{Y})$ is another operation that converts a matrix  $\mathbf{Y} \in \mathbb{R}^{m\times n}$ into a vector by stacking its columns. 

The following two sub-sections describe our proposed Kronecker modules. See Appendix \ref{sec:ab} for the ablation experiments justifying our design choices. These include studying the effect of initialization, the use of scale factor and non-linearity, as well as the position of the Kronecker module (parallel vs sequential). 
    
\subsection{KronA}
Figure \ref{fig:1}.a shows the structure of a LoRA module where $\mathbf{A}$ is the down projection and $\mathbf{B}$ is the up projection. To modify this module into KronA, the normal product is replaced by the Kronecker product. Also, LoRA projections are replaced with the Kronecker factors (See Figure \ref{fig:1}.b and Appendix  \ref{sec:app_a}). Equation \ref{eq:krona} shows how the output is generated when KronA is applied. $\mathbf{A}_k$ and $\mathbf{B}_k$ are the Kronecker factors that replaced the LoRA projections. Similar to LoRA, our KronA has a fixed scale factor, $s$, which is a hyperparameter.
\vspace{-2pt}
\begin{equation}
\label{eq:krona}
\mathbf{Y}= \mathbf{X}\mathbf{W} + s\mathbf{X}[\mathbf{A}_k\otimes\mathbf{B}_k]
\end{equation}
KronA modules are applied in parallel to the weight matrices in a PLM during the tuning phase. Once fine-tuned, the Kronecker factors are multiplied, then scaled and merged to the original PLM weight matrix (Equation \ref{eq:krona-merge}). Therefore, similar to LoRA, KronA does not increase the inference time. 
\vspace{-5pt}
\begin{equation}
\label{eq:krona-merge}
\mathbf{W}_{\text{tuned}}=\mathbf{W}+s[\mathbf{A}_k\otimes\mathbf{B}_k]
\end{equation}
For initialization of the Kronecker factors, we observed that initializing one of the factors with zero significantly improves the results compared to initializing both factors using the Normal distribution. See Appendix \ref{sec:ab} for more details.

\subsection{$\text{KronA}^\text{B}$ and $\text{KronA}^\text{B}_\text{res}$}
Inspired by the promising performance of the PA method, we also investigate KronA module when used in parallel to the FNN block and call it KronA$^\text{B}$. The B superscript in the name means that this module is applied to the PLM \underline{b}locks, opposed to KronA which is applied to the PLM weight matrices. Note that similar to the PA method, the non-linearity in the FFN block does not allow our KronA module to be merged into the pre-trained model after fine-tuning. This imposes an increase in the inference time and computations. Equation \ref{eq:kronam} shows how KronA$^\text{B}$ works in parallel to the FFN block.
\begin{equation}
\label{eq:kronam}
\mathbf{Y} = \text{FFN}(\mathbf{X}) + s\mathbf{X}[\mathbf{A}_
k\otimes \mathbf{B}_k]
\end{equation}

To further improve the representation power, we incorporate a scaled residual connection inside the $\text{KronA}^\text{B}$ module to develop $\text{KronA}^\text{B}_\text{res}$. Scale of the residual connection ($s_{res}$) is initialized with one and tuned during the fine-tuning. Equation \ref{eq:kronamr} shows how the $\text{KronA}^\text{B}_\text{res}$ works in parallel to the FFN. Also, Figure \ref{fig:1}.c and \ref{fig:1}.d show the structure of the PA module and our $\text{KronA}^\text{B}_\text{res}$, respectively.
\begin{equation}
\label{eq:kronamr}
\mathbf{Y} =  \text{FFN}(\mathbf{X}) + s\mathbf{X}[\mathbf{A}_K\otimes \mathbf{B}_K] + s_{res}\mathbf{X}
\end{equation}

\begin{table*}[tp!]
  \centering
  \resizebox{\textwidth}{!}{%
  \begin{tabular}{l|c|cccccccc|c}
    \toprule
      Method & Params & CoLA & RTE & MRPC & SST-2 & STS-B & MNLI & QNLI & QQP & Avg \\
    \toprule
        FT & 100 & 63.37 & 74.82 & 92.73 & 93.58 & 90.07 & 86.16 & 92.77 & 91.74 & 85.65  \\ 
    \midrule
        BitFit & 0.12 & 58.19 & 68.34 & 92.58 & 94.61 & 90.69 & 85.73 & 92.91 & 90.33 & 84.17 \\
        Adapter & 0.07 & 64.66 & 71.94 & 91.27 & \textbf{94.84} & 90.49 & 85.91 & 92.97 & 90.35 & 85.30 \\
        LoRA & 0.07 & 64.76 & 74.10 & 92.10 & 93.92 & 91.21 & 86.08 & 92.97 & \textbf{90.68} & 85.73 \\
        Compacter & 0.07 &  64.42 & 76.26 & 91.52 & 93.92 & 91.04 & 86.14 & 92.93 & 90.36 & 85.82 \\
        PA & 0.06 &  64.80 & 74.10 & \textbf{93.20} & 94.04 & 91.10 & 86.24 & 93.12 & 90.30 & 85.86 \\
        \midrule
        KronA & 0.07 & 63.27 & \textbf{77.70} & 92.52 & 94.26 & 91.30 & \textbf{86.34} & 93.15 & 90.57 & \textbf{86.14} \\
        $\text{KronA}^\text{B}$ & $0.07^*$ & 65.74 & 75.54 & 92.78 & 94.72 & \textbf{91.41} & 86.22 & 93.19 & \textbf{90.68} & \textbf{86.28} \\
        $\text{KronA}^\text{B}_\text{res}$ & $0.07^*$ & \textbf{66.73} & 76.98 & 93.15 & 94.38 & 91.35 & 86.20 & \textbf{93.21} & 90.57 & \textbf{86.57} \\
    \bottomrule
  \end{tabular}%
  }
  \caption{This table shows performance of the methods on the GLUE. $*$ shows that number of parameters might be a slightly different depending on the choice of one or two bias vectors in our KronA module.}
  \label{tab:1}
\end{table*}

\begin{table*}[tp!]
  \centering
  \resizebox{\textwidth}{!}{%
  \begin{tabular}{l|ccccccccc}
    \toprule
      Method &  FT & LoRA & KronA & BitFit & Adapter & PA & Compacter & $\text{KronA}^\text{B}$ & $\text{KronA}^\text{B}_\text{res}$ \\
    \toprule
      Inference Latency($\%$) & 100 & 100 & 100 & 100 & 146 & 113 & 181 & 127 & 136 \\ 
      \midrule
      Training Time($\%$) & 100 & 72 & 75 & 64 & 73 & 71 & 79 & 74 & 81 \\
    \bottomrule
  \end{tabular}%
  }
  \caption{In this table, the first row shows the normalized latency of the methods in the inference phase, and the second row lists the average normalized training time of the methods on the GLUE tasks.}
  \label{tab:2}
\end{table*}

\section{Results and Discussion}
\label{sec:exp}
\subsection{GLUE Results}
Table \ref{tab:1} shows\footnote{Details regarding the hyperparameters and experimental setups are mentioned in Appendix \ref{sec:app_H}.} the GLUE score of our proposed methods compared to other baselines when applied to the T5 model \cite{raffel2019exploring}. As the results show, KronA and KronA$^\text{B}$ outperform LoRA and PA as their low-rank counterparts, respectively. Also, all of our proposed modules outperform other state-of-the-art baselines on average and most of the GLUE tasks. In addition, $\text{KronA}^\text{B}_\text{res}$ which benefits from an extra learnable residual connection achieves remarkably better results. 


\subsection{Inference and Training Time}
Table \ref{tab:2} shows the normalized inference delay for the discussed methods. KronA, LoRA, FT, and Bitfit do not increase the inference latency since these techniques do not add any extra parameters or computations to the model during the inference stage. Although $\text{KronA}^\text{B}$ is significantly faster than Compacter and Adapter, it is slower than PA which is as we expected; calculation of the Kronecker product is generally slower than normal matrix multiplication. Also, adding the learnable residual connection increases the latency.

Normalized training time averaged over the GLUE tasks for each technique is shown in Table \ref{tab:2}. Based on these results, the significant improvement in the accuracy of the proposed KronA module and its variants is at the expense of a slight increase in the training time compared to the low-rank counterparts like LoRA, PA, and Adapter. However, the training time increase is not remarkable and KronA modules are still significantly faster than FT.
\vspace{-5pt}
\section{Conclusion and Future Directions}
\label{sec:con}
In this work, we developed Kronecker-based adapters by replacing the low-rank projections of the state-of-the-art PET methods with the Kronecker product. In addition to a comparison of the training and inference time, we evaluated our proposed adapter for fine-tuning T5 on the GLUE benchmark to show its superiority over the state-of-the-art baselines.




\bibliography{anthology,custom}
\bibliographystyle{acl_natbib}

\appendix
\section{Kronecker Factors vs Down and Up Projections}
\label{sec:app_a}
Table \ref{tab:replacement-shapes} shows details about the Kronecker factors that replaced the projections in the LoRA module. Note that $r<\frac{d_h}{2}$ to reduce the number of trainable parameters compared to the original PLM matrix.  
\begin{table*}[bp!]
    \centering
    \resizebox{\textwidth}{!}{%
    \begin{tabular}{c|cccccc}
    \toprule
        Module Name & Factor Name & Symbol & Shape & Parameters & Module Parameters & Constraint \\
        \midrule
        \multirow{2}{*}{KronA} & Kronecker Factor & $\mathbf{A}_k$ & $a_1 \times a_2$ & $a_1a_2$ & \multirow{2}{*}{$a_1a_2+b_1b_2$} & \multirow{2}{*}{$a_1b_1=a_2b_2=d_h$}\\
        & Kronecker Factor & $\mathbf{B}_k$ & $b_1 \times b_2$ & $b_1b_2$ & &  \\
        \midrule
        \multirow{2}{*}{LoRA} & Down Projection & $\mathbf{A}$ & $d_h \times r$ & $d_hr$ & \multirow{2}{*}{$2d_hr$} & \multirow{2}{*}{$r<\frac{d_h}{2}$}\\
        & Up Projection & $\mathbf{B}$ & $r \times d_h$ & $d_hr$ & \\
    \bottomrule
    \end{tabular}%
    }
    \caption{This table compares some details of Kronecker factors with LoRA projections.}
    \label{tab:replacement-shapes}
\end{table*}
\begin{table*}[tp!]
  \centering
  \begin{tabular}{c|cccccccc|c}
    \toprule
        Init Method & CoLA & RTE & MRPC & SST2 & STSB & MNLI & QNLI & QQP & Avg \\
    \toprule
         $\mathbf{A}_K$,$\mathbf{B}_K$ $\thicksim$ Normal & 63.36 & 66.91 & 91.69 & 91.97 & 90.46 & 86.03 & 92.33 & 90.19 & 84.12 \\ 
        $\mathbf{A}_K$ $\thicksim$ KU ,$\mathbf{B}_K$=0 &  63.27 & 77.70 & 92.52 & 94.04 & 91.26 & 86.03 & 93.13 & 90.57 & 86.06 \\ 
    \bottomrule
  \end{tabular}
  \caption{This table shows performance of KronA on GLUE using different initialization options for the Kronecker factors.}
  \label{tab:8}
\end{table*}

\section{Ablation Study}
\label{sec:ab}
\subsection{KronA Initialization}
\label{sec:init}
Our empirical results show that initialization of the Kronecker factors affects the performance of the KronA module. Table \ref{tab:8} shows the performance of two investigated strategies for KronA initialization. We observe that by initializing one of the Kronecker factors from the Kaiming-uniform ($a=\sqrt{5}$) distribution and the other one with zero, KronA module performs significantly better than initializing both of the factors from the Normal ($\mu=0, \sigma=\frac{1}{\sqrt{d_h}}$, where $d_h$ is the embedding dimension) distribution.
\begin{table*}[tp!]
  \centering
  \begin{tabular}{c|cccccccc|c}
    \toprule
      Modification & CoLA & RTE & MRPC & SST-2 & STS-B & MNLI & QNLI & QQP & Avg \\
    \toprule
        Sequential & 15.26 & 53.28 & 86.39 & 87.38 & 83.78 & 74.70 & 84.29 & 86.63 & 71.46 \\
        Parallel & 58.17 & 69.78 & 91.58 & 93.81 & 90.86 & 85.68 & 93.35 & 90.14 & 84.17 \\
        Parallel+Scale (PS) & 62.27 & 70.50 & 91.58 & 94.04 & 91.01 & 86.16 & 93.39 & 90.61 & 84.94 \\
        PS+SiLU & 62.74 & 69.78 & 91.89 & 94.15 & 90.97 & 85.98 & 93.30 & 90.15 & 84.87 \\
        PS only on FFN & 63.74 & 72.66 & 92.20 & 94.72 & 90.98 & 85.98 & 93.12 & 90.68 & 85.51 \\
    \bottomrule
  \end{tabular}
  \caption{This table shows performance of $\text{KronA}^\text{B}$ after implementing step by step modifications on GLUE.}
  \label{tab:5}
\end{table*}

\begin{table*}[tp!]
  \centering
  \begin{tabular}{c|ccccc}
    \toprule
        nonlinear function & mish & relu & gelu & $\text{gelu}_{\text{new}}$ & silu \\
    \toprule
        QNLI Performance & 93.21 & 93.28 & 93.13 & 93.26 & 93.30 \\ 
    \bottomrule
  \end{tabular}
  \caption{This table shows performance of $\text{KronA}^\text{B}$ on QNLI using different non-linear functions.}
  \label{tab:6}
\end{table*}

\subsection{Step by Step Improvement of  $\text{KronA}^\text{B}$}
At first, $\text{KronA}^\text{B}$ was initialized similar to a normal adapter. It was sequentially inserted after both FFN and attention blocks and it did not  have scaling. Based on \cite{he2022towards}, we made some modifications to our module to improve its performance. 

Table \ref{tab:5} shows the results of our experiments. We observed that inserting $\text{KronA}^\text{B}$ modules in parallel to the PLM modules instead of inserting them sequentially, improves performance significantly. Also, adding a scaling factor to our module increases the GLUE score further. Moreover, adding two modules to each FFN instead of adding to both FFN and attention blocks resulted in a higher score. 

In addition, motivated by the presence of a non-linear function in PA and Adapter, we tested different non-linear functions between the two multiplications (by $\mathbf{A}_k^{\text{T}}$ and $\mathbf{B}_k$) in Equation \ref{eq:kronecker_equivalance}. As Table \ref{tab:6} shows, SiLU \cite{Elfwing2018SigmoidWeightedLU} is the best option among others, but based on Table \ref{tab:5}, adding SiLU decreases GLUE score of the $\text{KronA}^\text{B}$. Therefore, we removed the non-linearity from our module. 

\subsection{Learnable Residual Connection}
In the $\text{KronA}^\text{B}_\text{res}$ module, a residual connection that is multiplied by a learnable scale is added to the output of $\text{KronA}^\text{B}$. Also, we studied another scenario in which the residual is multiplied by $\text{Sigmoid(learnable scale)}$. This module is called $\text{KronA}^\text{B}_\text{sigres}$. We wanted to answer the question "Is it better to limit the residual scale between 0 and 1?". Our empirical results (Table \ref{tab:res}) show that by adding a Sigmoid function, the performance of the module drops and the latency increases. Therefore, the Sigmoid function was removed from our module.
\begin{table}
  \centering
  \begin{tabular}{c|c|c}
    \toprule
      Method & Avg Score & Latency  \\
     \midrule
      $\text{KronA}^\text{B}_\text{res}$ & \textbf{86.57} & 1 \\
      $\text{KronA}^\text{B}_\text{sigres}$ & 86.42 & 1.18\\
    \bottomrule
  \end{tabular}
  \caption{This table shows the effect of adding a Sigmoid function to the $\text{KronA}^\text{B}_\text{res}$ module. Avg Score is the averaged score on the GLUE tasks and Latency represents the relative training time.}
  \label{tab:res}
\end{table}

\section{Details of Measuring Training and Inference Time}
To measure the inference latency, a random dummy input with the batch size equal to one and the sequence length equal to ten is generated. Then, the dummy input is given to the model for 150 iterations to warm-up the GPU. Finally, the dummy input is fed to the model for 200 iterations and the required time to generate the output is measured, averaged, and recorded. This experiment is repeated three times and the average latency is reported. Finally, the reported latencies are normalized and shown in Table \ref{tab:2}. 

Table \ref{tab:train-time} shows the normalized training time for each technique on the GLUE tasks. All the experiments are done with the same number of epochs, batch size, number of GPUs, and gradient accumulation step.

\begin{table*}[tp!]
  \centering
  \begin{tabular}{l|cccccccc|c}
    \toprule
      Method & CoLA & RTE & MRPC & SST-2 & STS-B & MNLI & QNLI & QQP & Avg \\
    \toprule
        FT & 1 & 1 & 1 & 1 & 1 & 1 & 1 & 1 & 1  \\
    \midrule
        BitFit & 0.58 & 0.64 & 0.65 & 0.66 & 0.66 & 0.65 & 0.64 & 0.62 & 0.64 \\
        Adapter & 0.82 & 0.71 & 0.72 & 0.78 & 0.76 & 0.72 & 0.69 & 0.69 & 0.73 \\
        LoRA &  0.79 & 0.69 & 0.7 & 0.76 & 0.72 & 0.7 & 0.68 & 0.68 & 0.72 \\
        KronA &  0.8 & 0.72 & 0.75 & 0.81 & 0.77 & 0.74 & 0.73 & 0.73 & 0.75 \\
        Compacter & 0.88 & 0.74 & 0.78 & 0.86 & 0.81 & 0.75 & 0.74 & 0.76 &  0.79 \\
        PA & 0.7 & 0.78 & 0.81 & 0.73 & 0.7 & 0.67 & 0.66 & 0.65 & 0.71 \\
        $\text{KronA}^\text{B}$ & 0.84 & 0.7 & 0.72 & 0.79 & 0.75 & 0.72 & 0.7 & 0.71 & 0.74 \\
        $\text{KronA}^\text{B}_\text{res}$ & 0.91 & 0.85 & 0.81 & 0.91 & 0.76 & 0.78 & 0.73 & 0.74 & 0.81 \\
    \bottomrule
  \end{tabular}
  \caption{This table shows the normalized training time of methods on the GLUE tasks.}
  \label{tab:train-time}
\end{table*}
\begin{table}[tp!]
  \centering
  \begin{tabular}{c|c}
    \toprule
      Shape of $\mathbf{A}_k$ & MNLI (Accuracy)  \\
     \midrule
      (48, 16) & 86.50 \\
      (32, 24) & 86.31 \\
      (3, 256) &  86.16 \\
      (24, 32) &  86.40 \\
      (2, 384) & \textbf{86.63} \\
      (192, 4) & 86.46 \\
      (12, 64) & 86.56 \\
    \bottomrule
  \end{tabular}
  \caption{This table shows the performance of the tested options for $\mathbf{A}_k$ in KronA on MNLI. Note that for each option, the shape of the corresponding $\mathbf{B}_k$ is in the reversed order of $\mathbf{A}_k$.}
  \label{tab:shape}
\end{table}
\begin{table*}[tp!]
    \centering
    \resizebox{\textwidth}{!}{%
    \begin{tabular}{c|ccccccccc}
        \toprule
        Method & FT & BitFit & Adapter & Compacter & LoRA & PA & KronA & $\text{KronA}^\text{B}$ & $\text{KronA}^\text{B}_\text{res}$ \\
        \midrule
        Block & Entire Model & Biases & FFN \& Attention & FFN \& Attention & Query \& Value & FFN & Query \& Value & FFN & FFN \\
        \bottomrule
    \end{tabular}%
    }
    \caption{This table shows the PLM block each method has been applied to.}
    \label{tab:module-place}
\end{table*}

\section{Experimental Setups and Hyperparameters}
\label{sec:app_H}
\subsection{Datasets}
We used the GLUE \cite{wang-etal-2018-glue} benchmark to evaluate our methods compared to the baselines. This benchmark covers a variety of tasks including natural language inference (MNLI, RTE, QNLI), linguistic
acceptability (CoLA), similarity and paraphrasing (MRPC, QQP), and sentiment classification (SST-2). The original test set of GLUE is not published, so similar to \cite{karimi2021compacter, zhang2022hyperpelt}, we generated our test sets from the evaluation and the training data. For the small datasets (CoLA, RTE, MRPC, and STSB), we use half of the  task dev set for evaluation and the other half as the test set. For the rest of the GLUE tasks with larger datasets, we take 1K samples out of the train set and use it as our test set. The reported evaluation metric for CoLA, MRPC, STS-B, is Matthew correlation coefficient, F1, and the average of Pearson/Spearman correlations, respectively. Accuracy is used for the other tasks.
\subsection{Experimental Setup}
All experiments were performed on one NVIDIA Tesla V100. We used PyTorch and Hugging Face Transformers library \cite{wolf2019huggingface} for our experiments. To re-implement \href{https://github.com/microsoft/LoRA}{LoRA} and \href{https://github.com/jxhe/unify-parameter-efficient-tuning}{PA}, we used their publicly available code. For the experiments on the Compacter, BitFit, FT and Adpater, we used the \href{https://github.com/rabeehk/compacter}{Compacter}'s official code. The Backbone model for this work is $\text{T5}_{\text{base}}$ \cite{raffel2019exploring}.

The size of the trainable parameters for all of the methods is set roughly equal to have a fair comparison. However, for the BitFit tuning we could not match the trainable parameters since all of the biases are trainable. 

Given the number of trainable parameters, we have several choices for the shapes of Kronecker factors. For KronA, We tested some of the options and selected the option with the best results.

Table \ref{tab:module-place} shows the specific PLM block on which each technique is applied.

$\text{KronA}^\text{B}$ and $\text{KronA}^\text{B}_{\text{res}}$ modules can have one or two biases. We selected the number of biases that maximized the score on each task. 

\subsection{Hyperparameters}
Since we wanted to ignore the effect of the scaling factor when comparing LoRA and KronA, the scaling factor for these two modules is set to one in all of the experiments. 

For FT, BitFit, Compacter, and Adapter experiments, we used the hyperparameters that are mentioned in \cite{karimi2021compacter}. However, we changed the learning rate and the rank of modules to match the desired number of trainable parameters in the Adapter experiments.

The rank of LoRA and PA is set to one and two, respectively. For the KronA modules, the shape of the Kronecker factors are selected based on the best dev results among different options for the shapes. Due to time and resource limitations, we did not tune the shape of the Kronecker factors for $\text{KronA}^\text{B}$ and $\text{KronA}^\text{B}_\text{res}$. 

All of the other hyperparameters are set based on \cite{karimi2021compacter}, expect for the learning rate and scaling factor which are tuned based on the best dev results. All of the methods are trained for 20 epochs and the checkpoint that achieves the best performance on dev set is reported as the final model. Tables \ref{tab:pet:glue:ft}, \ref{tab:pet:glue:bitfit}, \ref{tab:pet:glue:adapter}, \ref{tab:pet:glue:compacter}, \ref{tab:pet:glue:lora}, \ref{tab:pet:glue:pa}, \ref{tab:pet:glue:krona}, \ref{tab:pet:glue:kronam} and \ref{tab:pet:glue:kronamr} show the tuned hyperparameters for each method on the GLUE tasks.
\begin{table*}
  \centering
  \begin{tabular}{cccccc}
    \toprule
    \multicolumn{6}{c}{\textbf{FT hyperparameters}} \\
    \toprule
    Task & learning rate & batch size & warmup steps & source sentence length & epoch \\
    \midrule
    GLUE & 3e-4 & 100 & 500 & 128 & 20 \\
    \bottomrule
  \end{tabular}
  \caption{This table shows the hyperparameters used for FT experiments on the GLUE tasks.}
  \label{tab:pet:glue:ft}
\end{table*}
\begin{table*}[tp!]
  \centering
  \begin{tabular}{cccccc}
    \toprule
    \multicolumn{6}{c}{\textbf{BitFit hyperparameters}} \\
    \toprule
    Task & learning rate & batch size & warmup steps & source sentence length & epoch \\
    \midrule
    GLUE & 3e-4 & 100 & 500 & 128 & 20 \\
    \bottomrule
  \end{tabular}
  \caption{This table shows the hyperparameters used for BitFit experiments on the GLUE tasks.}
  \label{tab:pet:glue:bitfit}
\end{table*}

\begin{table*}[tp!]
  \centering
  \begin{tabular}{ccccc}
    \toprule
    \multicolumn{5}{c}{\textbf{Adapter hyperparameters}} \\
    \toprule
    Task & learning rate & batch size & task reduction factor & epoch \\
    \midrule
    GLUE & 3e-3 & 100 & 32 & 20 \\
    \bottomrule
  \end{tabular}
  \caption{This table shows the hyperparameters used for Adapter experiments on the GLUE tasks.}
  \label{tab:pet:glue:adapter}
\end{table*}

\begin{table*}[tp!]
  \centering
  \resizebox{\textwidth}{!}{%
  \begin{tabular}{cccccc}
    \toprule
    \multicolumn{6}{c}{\textbf{Compacter hyperparameters}} \\
    \toprule
    Task & learning rate & batch size & hypercomplex division & task reduction factor & epoch \\
    \midrule
    GLUE & 3e-3 & 100 & 4 & 32 & 20 \\
    \bottomrule
  \end{tabular}%
  }
  \caption{This table shows the hyperparameters used for Compacter experiments on the GLUE tasks.}
  \label{tab:pet:glue:compacter}
\end{table*}
\begin{table*}[b!]
  \centering
  \begin{tabular}{cccccc}
    \toprule
    \multicolumn{6}{c}{\textbf{LoRA hyperparameters}} \\
    \toprule
    Task & learning rate & batch size & rank & $s$ & epoch \\
    \midrule
    GLUE & 1e-3 & 100 & 1 & 1 & 20 \\
    \bottomrule
  \end{tabular}
  \caption{This table shows the hyperparameters used for LoRA experiments on the GLUE tasks.}
  \label{tab:pet:glue:lora}
\end{table*}

\begin{table*}
  \centering
  \begin{tabular}{cccccc}
    \toprule
    \multicolumn{6}{c}{\textbf{PA hyperparameters}} \\
    \toprule
    Task & learning rate & batch size & rank & $s$ & epoch \\
    \midrule
    CoLA & 3e-3 & 100 & 2 & 16 & 20 \\
    \midrule
    RTE & 5e-3 & 100 & 2 & 16 & 20 \\
    \midrule
    MRPC & 5e-3 & 100 & 2 & 16 & 20 \\
    \midrule
    SST-2 & 1e-3 & 100 & 2 & 16 & 20 \\
    \midrule
    SSTS-B & 1e-3 & 100 & 2 & 16 & 20 \\
    \midrule
    MNLI & 1e-3 & 100 & 2 & 16 & 20 \\
    \midrule
    QNLI & 1e-3 & 100 & 2 & 16 & 20 \\
    \midrule
    QQP & 1e-3 & 100 & 2 & 16 & 20 \\
    \bottomrule
  \end{tabular}
  \caption{This table shows the hyperparameters used for PA experiments on the GLUE tasks.}
  \label{tab:pet:glue:pa}
\end{table*}

\begin{table*}
  \centering
  \begin{tabular}{ccccccc}
    \toprule
    \multicolumn{7}{c}{\textbf{KronA hyperparameters}} \\
    \toprule
    Task & learning rate & batch size & $\mathbf{A}_k$ & $\mathbf{B}_k$ & $s$ & epoch \\
    \midrule
    CoLA & 1e-3 & 100 & (32,24) & (24,32) & 1 & 20 \\
    \midrule
    RTE & 2e-3 & 100 & (32,24) & (24,32) & 1 & 20 \\
    \midrule
    MRPC & 1e-3 & 100 & (32,24) & (24,32) & 1 & 20 \\
    \midrule
    SST-2 & 1e-3 & 100 & (24,32) & (32,24) & 1 & 20 \\
    \midrule
    SSTS-B & 1e-3 & 100 & (2,384) & (384,2) & 1 & 20 \\
    \midrule
    MNLI & 1e-3 & 100 & (2,384) & (384,2) & 1 & 20 \\
    \midrule
    QNLI & 1e-3 & 100 & (3,256) & (256,3) & 1 & 20 \\
    \midrule
    QQP & 1e-3 & 100 & (24,32) & (32,24) & 1 & 20 \\
    \bottomrule
  \end{tabular}
  \caption{This table shows the hyperparameters used for KronA experiments on the GLUE tasks.}
  \label{tab:pet:glue:krona}
\end{table*}
\begin{table*}
  \centering
  \begin{tabular}{cccccccc}
    \toprule
    \multicolumn{8}{c}{\textbf{$\text{KronA}^\text{B}$ hyperparameters}} \\
    \toprule
    Task & learning rate & batch size & $\mathbf{A}_k$ & $\mathbf{B}_k$ & $s$ & module bias & epoch \\
    \midrule
    CoLA & 1e-3 & 100 & (32,24) & (24,32) & 16 & 2 & 20 \\
    \midrule
    RTE & 5e-3 & 100 & (32,24) & (24,32) & 16 & 1 & 20 \\
    \midrule
    MRPC & 5e-3 & 100 & (32,24) & (24,32) & 16 & 1 & 20 \\
    \midrule
    SST-2 & 1e-3 & 100 & (32,24) & (24,32) & 4 & 1 & 20 \\
    \midrule
    SSTS-B & 1e-3 & 100 & (32,24) & (32,24) & 16 & 1 & 20 \\
    \midrule
    MNLI & 1e-3 & 100 & (32,24) & (24,32) & 4 & 2 & 20 \\
    \midrule
    QNLI & 1e-3 & 100 & (32,24) & (24,32) & 4 & 1 & 20 \\
    \midrule
    QQP & 1e-3 & 100 & (32,24) & (24,32) & 4 & 1 & 20 \\
    \bottomrule
  \end{tabular}
  \caption{This table shows the hyperparameters used for $\text{KronA}^\text{B}$ experiments on the GLUE tasks.}
  \label{tab:pet:glue:kronam}
\end{table*}
\begin{table*}
  \centering
  \begin{tabular}{cccccccc}
    \toprule
    \multicolumn{8}{c}{\textbf{$\text{KronA}^\text{B}_\text{res}$ hyperparameters}} \\
    \toprule
    Task & learning rate & batch size & $\mathbf{A}_k$ & $\mathbf{B}_k$ & $s$ & module bias & epoch \\
    \midrule
    CoLA & 1e-3 & 100 & (32,24) & (24,32) & 16 & 2 & 20 \\
    \midrule
    RTE & 5e-3 & 100 & (32,24) & (24,32) & 16 & 2 & 20 \\
    \midrule
    MRPC & 5e-3 & 100 & (32,24) & (24,32) & 16 & 2 & 20 \\
    \midrule
    SST-2 & 1e-3 & 100 & (32,24) & (24,32) & 16 & 1 & 20 \\
    \midrule
    SSTS-B & 9e-4 & 100 & (32,24) & (32,24) & 16 & 1 & 20 \\
    \midrule
    MNLI & 1e-3 & 100 & (32,24) & (24,32) & 16 & 1 & 20 \\
    \midrule
    QNLI & 1e-3 & 100 & (32,24) & (24,32) & 4 & 2 & 20 \\
    \midrule
    QQP & 1e-3 & 100 & (32,24) & (24,32) & 16 & 1 & 20 \\
    \bottomrule
  \end{tabular}
  \caption{This table shows the hyperparameters used for $\text{KronA}^\text{B}_\text{res}$ experiments on the GLUE tasks.}
  \label{tab:pet:glue:kronamr}
\end{table*}

\end{document}